\documentclass[number,preprint,review,12pt]{elsarticle}

\usepackage{amssymb}
\usepackage{amsmath}
\usepackage{url} 

\usepackage{natbib}
\usepackage{subfigure}
\usepackage[table]{xcolor}
\usepackage{pdflscape}
\usepackage{geometry}
\usepackage{svg}
\definecolor{darkred}{RGB}{139,0,0}

\usepackage[normalem]{ulem}
\usepackage{ifthen}

\newcommand*{\ano}{0} 

\newcommand*{\authorN}[2]{\ifthenelse{\equal{\ano}{1}}{#2}{#1}}
\newcommand*{\github}[1]{\ifthenelse{\equal{\ano}{1}}{\textit{Github removed for anonymization}}{\url{#1}}}
\newcommand*{\by}[2]{\ifthenelse{\equal{\ano}{1}}{\textit{\subsection{\textbf{By:} #2}}}{\subsection{\textbf{By:} #1}}}
\newcommand*{\methodby}[1]{\ifthenelse{\equal{\ano}{1}}{\textit{\textbf{By:} Authors removed for anonymisation}}{\textbf{By:} #1}}

\journal{Computers & Graphics}

\begin{document}

\begin{frontmatter}



\title{SHREC 2025: Protein Surface Shape Retrieval including Electrostatic potential}


\author[CNAM]{Taher Yacoub\corref{cor1}\fnref{fn1,fn2}}
\emailauthor{taher.yacoub@lecnam.net}{Taher Yacoub}

\author[CNAM]{Camille Depenveiller\corref{cor1}\fnref{fn1,fn2}}
\emailauthor{camille.depenveiller@lecnam.net}{Camille Depenveiller}

\author[TSUYAMA]{Atsushi Tatsuma}
\emailauthor{tatsuma@tsuyama-ct.ac.jp}{Atsushi Tatsuma}

\author[CONTACT]{Tin Barisin}
\emailauthor{tin.barisin@contact-software.com}{Tin Barisin}
\author[CONTACT]{Eugen Rusakov}
\emailauthor{eugen.rusakov@contact-software.com}{Eugen Rusakov}
\author[CONTACT]{Udo Göbel}
\emailauthor{udo.goebel@contact-software.com}{Udo Göbel}

\author[CHANGSHA]{Yuxu Peng}
\emailauthor{pengyuxupjl@csust.edu.cn}{Yuxu Peng}
\author[CHANGSHA]{Shiqiang Deng}
\emailauthor{sqdeng@stu.csust.edu.cn}{Shiqiang Deng}

\author[PURDUE1]{Yuki Kagaya}
\emailauthor{ykagaya@purdue.edu}{Yuki Kagaya}
\author[PURDUE2]{Joon Hong Park}
\emailauthor{park1617@purdue.edu}{Joon Hong Park}
\author[PURDUE1,PURDUE2]{Daisuke Kihara}
\emailauthor{dkihara@purdue.edu}{Daisuke Kihara}

\author[IMATI]{Marco Guerra}
\emailauthor{marco.guerra@ge.imati.cnr.it}{Marco Guerra}
\author[IMATI]{Giorgio Palmieri}
\emailauthor{giorgio.palmieri@ge.imati.cnr.it}{Giorgio Palmieri}
\author[IMATI]{Andrea Ranieri}
\emailauthor{andrea.ranieri@ge.imati.cnr.it}{Andrea Ranieri}
\author[IMATI]{Ulderico Fugacci}
\emailauthor{ulderico.fugacci@ge.imati.cnr.it}{Ulderico Fugacci}
\author[IMATI]{Silvia Biasotti}
\emailauthor{silvia.biasotti@ge.imati.cnr.it}{Silvia Biasotti}

\author[DEVINCI]{Ruiwen He}
\emailauthor{ruiwen.he@devinci.fr}{Ruiwen He}
\author[IMT]{Halim Benhabiles}
\emailauthor{halim.benhabiles@imt-nord-europe.fr}{Halim Benhabiles}
\author[ESIGELEC]{Adnane Cabani}
\emailauthor{adnane.cabani@esigelec.fr}{Adnane Cabani}
\author[IRIMAS]{Karim Hammoudi}
\emailauthor{karim.hammoudi@uha.fr}{Karim Hammoudi}

\author[YUNNAN]{Haotian Li}
\emailauthor{2424420011@ynnu.edu.cn}{Haotian Li}
\author[NYU]{Hao Huang}
\emailauthor{hh1811@nyu.edu}{Hao Huang}
\author[YUNNAN]{Chunyan Li}
\emailauthor{lchy@ynnu.edu.cn}{Chunyan Li}

\author[QUEENS]{Alireza Tehrani}
\emailauthor{19at27@queensu.ca}{Alireza Tehrani}
\author[QUEENS]{Fanwang Meng}
\emailauthor{fanwang.meng@queensu.ca}{Fanwang Meng}
\author[QUEENS]{Farnaz Heidar-Zadeh}
\emailauthor{farnaz.heidarzadeh@queensu.ca}{Farnaz Heidar-Zadeh}

\author[VNUHCM,VNU]{Tuan-Anh Yang}
\emailauthor{ytanh21@apcs.fitus.edu.vn}{Tuan-Anh Yang}

\author[CNAM,IUF]{Matthieu Montes\corref{cor1}\fnref{fn2}}
\emailauthor{matthieu.montes@cnam.fr}{Matthieu Montes}

\affiliation[CNAM]{organization={Laboratoire GBCM, EA7528, Conservatoire Nationale des Arts et Métiers}, city={Paris}, country={France}}
\affiliation[IUF]{organization={Institut Universitaire de France}, city={Paris}, country={France}}
\affiliation[TSUYAMA]{organization={National Institute of Technology, Tsuyama College}, city={Tsuyama}, country={Japan}}
\affiliation[CONTACT]{organization={CONTACT AI Research, CONTACT Software GmbH}, city={Bremen}, country={Germany}}
\affiliation[CHANGSHA]{organization={Changsha University of Science and Technology}, city={Hunan}, country={China}}
\affiliation[PURDUE1]{organization={Department of Biological Sciences, Purdue University}, city={West Lafayette, IN}, country={USA}}
\affiliation[PURDUE2]{organization={Department of Computer Science, Purdue University}, city={West Lafayette, IN}, country={USA}}
\affiliation[IMATI]{organization={CNR-IMATI}, city={Genova}, country={Italy}}
\affiliation[DEVINCI]{organization={De Vinci Higher Education, De Vinci Research Center}, city={Paris}, country={France}}
\affiliation[IMT]{organization={IMT Nord Europe, Institut Mines-Télécom, Université de Lille, Center for Digital Systems}, city={Lille}, country={France}}
\affiliation[ESIGELEC]{organization={Université de Rouen Normandie, ESIGELEC, IRSEEM}, city={Rouen}, country={France}}
\affiliation[IRIMAS]{organization={IRIMAS, Université de Haute-Alsace}, city={Mulhouse}, country={France}}
\affiliation[YUNNAN]{organization={Yunnan Normal University}, city={Kunming, Yunnan}, country={China}}
\affiliation[NYU]{organization={New York University Abu Dhabi}, city={Abu Dhabi}, country={UAE}}
\affiliation[QUEENS]{organization={Department of Chemistry, Queen’s University}, city={Kingston, ON}, country={Canada}}
\affiliation[VNUHCM]{organization={University of Science, VNU-HCM}, city={Ho Chi Minh City}, country={Vietnam}}
\affiliation[VNU]{organization={Vietnam National University}, city={Ho Chi Minh City}, country={Vietnam}}

\fntext[fn1]{Equal contributions.}
\fntext[fn2]{Co-organizers of the track.}
\cortext[cor1]{Corresponding authors: Taher Yacoub, Camille Depenveiller, Matthieu Montes}

\begin{abstract}

This SHREC 2025 track dedicated to protein surface shape retrieval involved 9 participating teams. We evaluated the performance in retrieval of 15 proposed methods on a large dataset of 11,565 protein surfaces with calculated electrostatic potential (a key molecular surface descriptor). The performance in retrieval of the proposed methods was evaluated through different metrics (Accuracy, Balanced accuracy, F1 score, Precision and Recall). The best retrieval performance was achieved by the proposed methods that used the electrostatic potential complementary to molecular surface shape. This observation was also valid for classes with limited data which highlights the importance of taking into account additional molecular surface descriptors.

\end{abstract}




\end{frontmatter}



\section{Introduction}

Proteins are macromolecules involved in most biological processes. They are classified according to their evolutionary relationship based on their structure/fold (SCOPe and CATH)~\cite{chandonia2018scope,CATH}, or sequence similarity (Pfam)~\cite{PFAM}. Proteins interact through their molecular surface~\cite{connolly}, which is an abstraction of the underlying protein sequence, structure, and fold~\cite{wolfson04,dokholyan2009,gainza2019deciphering,machat21}. This abstraction can lead to the definition of protein surficial homologs: proteins that share a similar molecular surface shape. Even with low sequence or structure/fold similarity, remote protein surficial homologs can share \textit{in vivo} similar functions or partners involved in interaction~\cite{surfup2007, dokholyan2009,Sael2008fast,machat21,riahi2023_surfaceid}. For example, some viral proteins can mimic similar functions despite different structures~\cite{alcami2003}, or two proteins belonging to the same. CATH or SCOPe classification can be further refined by incorporating structural properties in order to better define or deduce biological functions, for example the subdivision of an enzyme into different subtypes~\cite{Tseng2012}. \\

A comparative study of shape retrieval methods in the context of proteins retrieval was done recently~\cite{machat21}, whose selected methods in this study can be considered as reference methods of these last years. Shape retrieval was tackled by different types of methods: (i) shape retrieval methods based on histograms summarizing global and/or local geometric descriptors~\cite{rusu2010fast}, (ii) shape retrieval methods based on spectral geometric descriptors~\cite{sirugue2024plo3s,reuter2006laplace}, (iii) shape retrieval methods based on molecular surface maps (projection of 3D protein into 2D map)~\cite{sirugue2024plo3s,papadakis2010panorama}, (iv) shape retrieval methods based on the moments of 3D Zernike descriptors~\cite{La2009}, and recently, (v) shape retrieval methods based on geometric machine/deep learning~\cite{mallet2024atomsurf, sharp2022diffusionnet,sverrisson2021dmasif,gainza2019deciphering,monti2017geometric,Qi2017}. Different SHREC tracks have been dedicated to protein surface shapes retrieval to evaluate these different methods~\cite{mavridis2010shrec,song2017protein,langenfeld2018shrec,langenfeld2019shrec,LANGENFELD2020189,langenfeld2021shrec,raffo2021shrec,gagliardi2022shrec,yacoub2024shrec}. Some of these SHREC tracks included the description in the dataset of molecular surface physicochemical properties in addition to the protein surface shape~\cite{langenfeld2021shrec,raffo2021shrec,gagliardi2022shrec}. A summary of the modalities and features (proteins, dataset, methods) of these tracks in comparison to this track is presented in Table \ref{tab:tracks}. \\

In the present SHREC track dedicated to protein shapes retrieval, we decided to evaluate the impact of including electrostatic potential, a key molecular surface descriptor~\cite{udock}, in the performance in retrieval of the 15 different methods proposed by the 9 participating teams (see Table \ref{tab:methods}). Comparatively to previous protein shape retrieval SHREC tracks, our track includes a much larger dataset, proteins with surficial electrostatic potential, more alternate conformations, and more homologous proteins by shape and sequence. A recent work~\cite{mallet2024atomsurf} highlighted the promise of surface-based learning, but also points out its current limitations which may be addressed by combining multiple representations, such as surface shape and graph-based representations. Our track was therefore configured to evaluate mainly ML/DL methods including training and test subsets. Out of the 15 methods proposed, only one was learning-free, allowing for a direct comparison with a non-learning-based approach. This learning-free method was evaluated in the previous tracks with good retrieval performances. 

The paper is organized as follows: Section~\ref{meth} describes the dataset structure and composition and the different methods used to evaluate the performance in retrieval of the proposed methods, Section~\ref{meth-ptcp} describes the participating teams proposed methods, Section~\ref{res} describes the results of each of the 15 proposed methods, Section~\ref{disc} and Section~\ref{conc} provide a discussion of the observed results and a conclusion about this SHREC track.

\clearpage
\newgeometry{margin=2.8cm}
\pagestyle{empty}
\begin{table}[htbp]
\centering
\noindent
\begin{minipage}[t]{1.5\textwidth} 
\begin{tabular}{|p{3.4cm}|p{6.0cm}|p{2.8cm}|p{2.0cm}|}
\hline
\textbf{Track} & \textbf{Protein Modalities}  & \textbf{Dataset Size (protein shapes)} & \textbf{Number of Methods/Runs} \\
\hline
SHREC 2010 (Mavridis \textit{et al.}) & From CATH database. Geometric similarity & 1,000 & 6 \\
\hline
SHREC 2017 (Song \textit{et al.}) & From PDB database. Geometric similarity & 5,854 & 6 \\
\hline
SHREC 2018 (Langenfeld \textit{et al.}) & Homologous protein domains. NMR including structural flexibility. Geometric property & 2,267 & 6 \\
\hline
SHREC 2019 (Langenfeld \textit{et al.}) & From SCOPe database. Focus on evolutionary relationships. NMR structures. Geometric property & 5,298 & 5 \\
\hline
SHREC 2020 (Langenfeld \textit{et al.}) & Same as SHREC 2019 but X-ray structures & 588 & 15 \\
\hline
SHREC 2021 (A) (Langenfeld \textit{et al.}) & From Pfam domains database. Geometriy + electrostatic properties & 554 & 5 \\
\hline
SHREC 2021 (B) (Raffo \textit{et al.}) & SHREC 2019 benchmark with physicochemical properties & 5,000 & 28 \\
\hline
SHREC 2022 (Gagliardi \textit{et al.}) & Local binding site recognition. From binding-MOAD database & 1,091 & 4 \\
\hline
SHREC 2024 (Yacoub \textit{et al.}) & Protein-protein complementary retrieval. From Protein-Protein Docking benchmark & 387 queries and 520 targets & 2 \\
\hline
\textbf{SHREC 2025 (Our track)} & From PDB100 and PDB90. Sequence, structural and shape homology. NMR, X-ray and cryo-EM structures. Geometry + electrostatic properties & 11,565 (9,244 in training set, 2,311 in test set) & 15 \\
\hline
\end{tabular}
\end{minipage}
\caption{Survey of the characteristics of current and previous tracks}
\label{tab:tracks}
\end{table}

\clearpage
\restoregeometry
\pagestyle{plain}

\section{Materials and Methods}\label{meth}

\subsection{Dataset}

\begin{figure}
    \centering
    \includegraphics[width=0.9\linewidth]{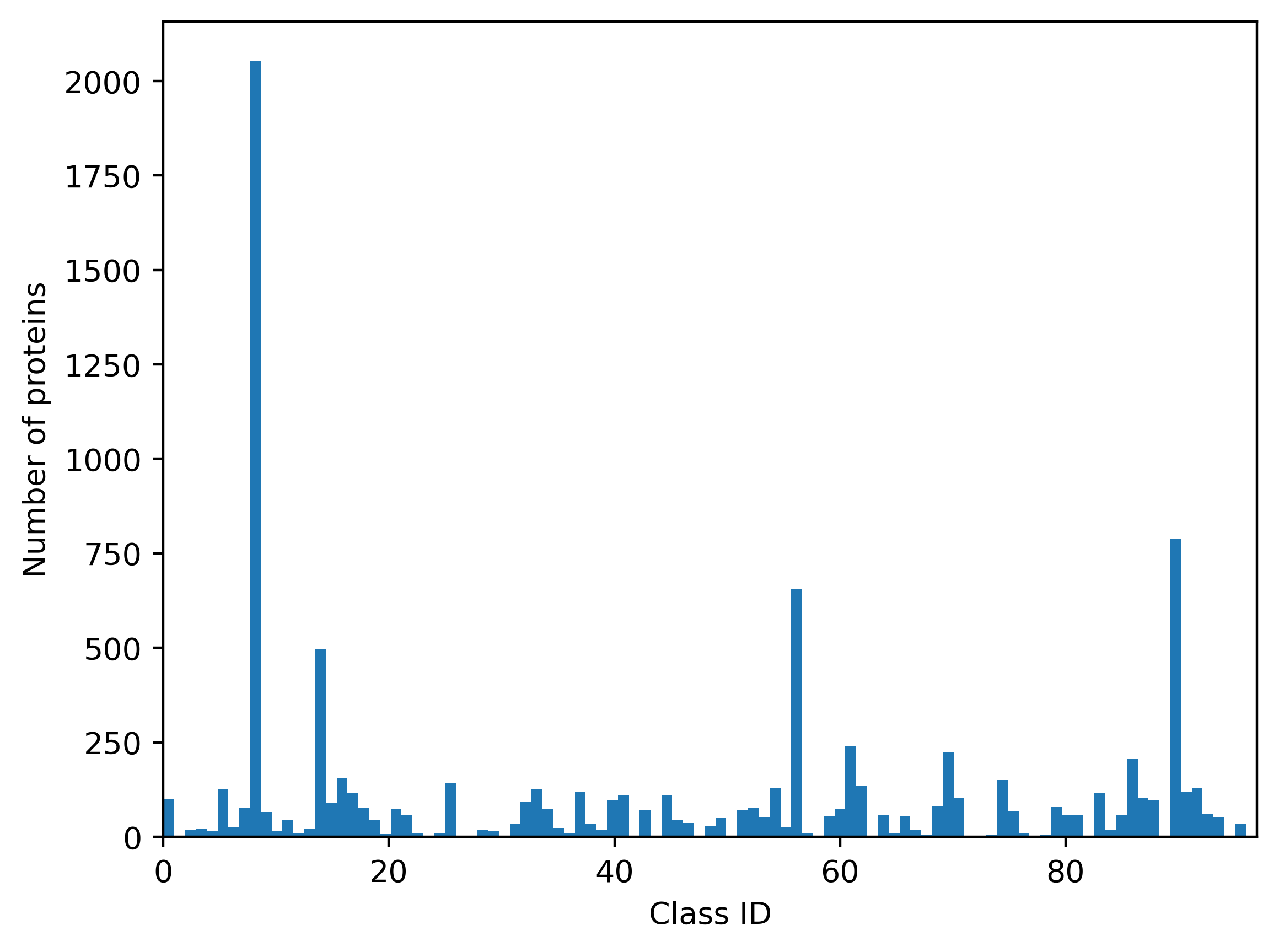}
    \caption{Class distribution in the train set}
    \label{fig:Train_distri}
\end{figure}

\begin{figure}
    \centering
    \includegraphics[width=0.9\linewidth]{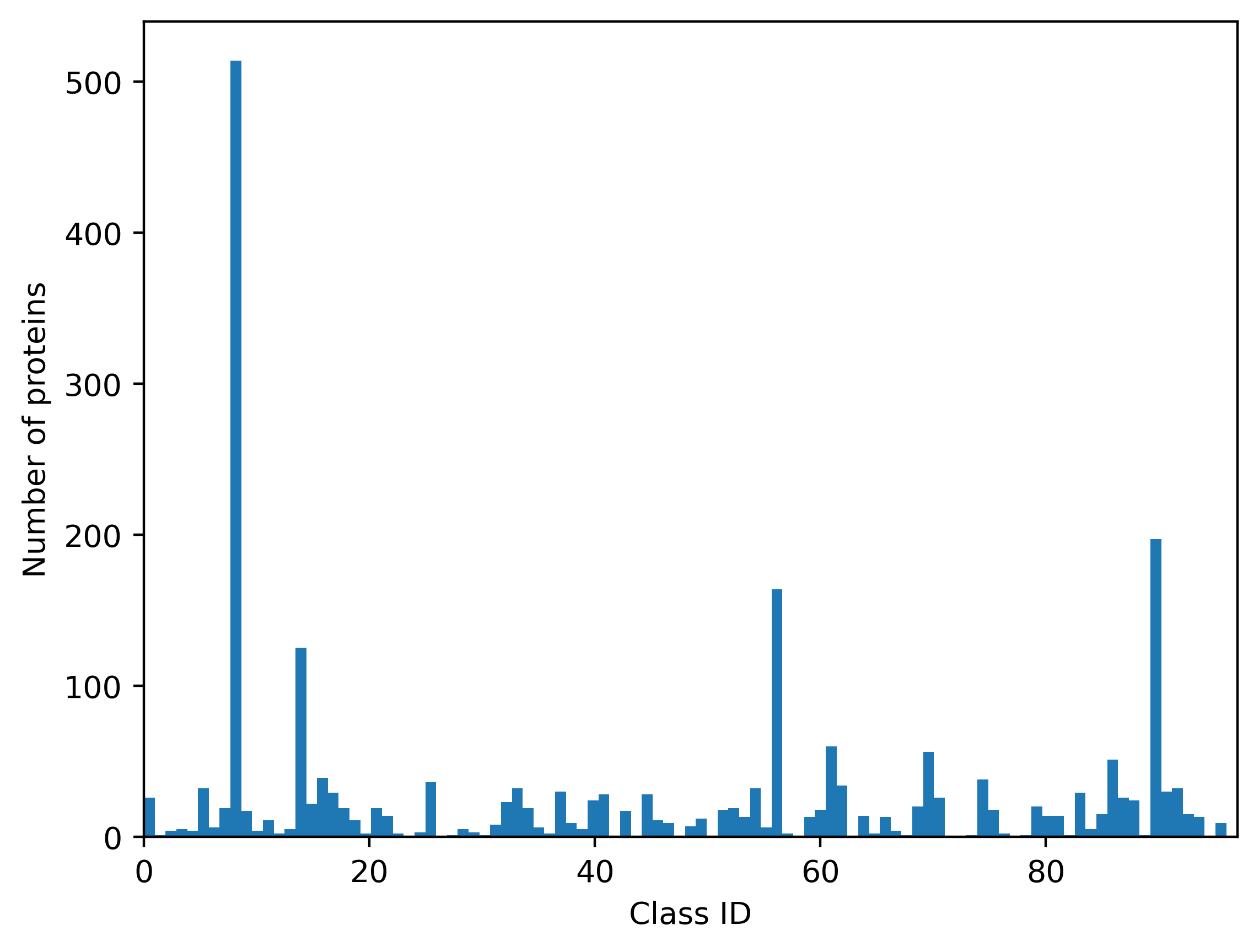}
    \caption{Class distribution in the test set}
    \label{fig:Test_distri}
\end{figure}

\noindent\textbf{Overview.} We proposed a dataset of 11,565 protein surfaces divided into 97 unbalanced classes that reflect current biological knowledge. This dataset was built on the basis of sequence and structure homology from the Protein Data Bank (RCSB) at 100\% and 90\% sequence identity. This selection allowed to select homologous proteins with similar electrostatic properties, independently to their biological function. All species were taken into consideration. The selected proteins are involved in a broad range of biological and biomedical functions. For example, cAMP beta-lactamase (e.g. 3GRJ, chain A, 358 residues) is a hydrolase involved in multidrug resistance; gamma-actin (e.g. 6EGT, chain A, 375 residues) is involved in non-syndromic hearing loss; HIV-1 reverse transcriptase in HIV infection process (3LP0, chain A, 563 residues). Other proteins correspond to subunits of ribosomes or viral capsids (e.g. 3J3Y, 231 residues). The size of proteins (with at least 50 residues) varies from 2,000 points to 90,000 points and was presented in Supplementary Data (Figure 5). A class is defined by a protein with different conformations (non-rigid deformations of protein surfaces). The smallest and largest (class 8) classes were composed of 2 and 2,568 protein surfaces, respectively. This dataset has been split into a training set and a test set in a 80/20 proportion using Scikit-learn tools~\cite{scikit-learn}. The training set included 9,244 protein surfaces with their corresponding ground truth. The test set included 2,311 protein surfaces without ground truth (see Figures~\ref{fig:Train_distri} and~\ref{fig:Test_distri}). The dataset can be downloaded at \authorN{\url{https://shrec2025.drugdesign.fr/}}{Anonymized link}. \\


\noindent\textbf{Protocol.} 
The dataset only included experimentally resolved structures from NMR, X-Ray crystallography or Cryo-EM structure available in the PDB~\cite{berman2000protein} with more than 50 residues.
To select homologous structures and alternate conformations, we used a procedure similar to~\cite{BarrioHernandez2023}. 
MMseqs2~\cite{Steinegger2017} clusters from the RCSB PDB Services (\url{www.rcsb.org/docs/programmatic-access/file-download-services}) at 100\% (PDB100) and 90\% (PDB90) sequence identity were used as a reference. Single- and multi-domain proteins were also included in the dataset. For PDBs multi-domain structures, they were already separated into individual domains (individual chains) by MMseqs2 RCSB PDB Services. The selection of structures was as follows: (i) only PDB100 sequence clusters with more than 20 experimental structures were selected; (ii) homologous proteins associated with each cluster of the PDB100 were selected from the PDB90 on the basis of 90-98\% sequence identity and a maximum sequence length difference less than or equal to 2\%. Based on these criteria, we assumed that high homology results in similar folds, which allowed us to create "homologous classes". The homology between classes is found in Supplementary Data Figure 1. 
To select alternate conformations, FoldSeek~\cite{vankempen2024} was used for each class using the "3Di+AA Gotoh-Smith-Waterman" algorithm. 

For each PDB structure, "PDB2PQR" tool computed charges and to get PQR files. Then, APBS~\cite{apbs} tool which includes TABI-PB (Treecode-Accelerated Boundary Integral Poisson-Boltzmann) solver and NanoShaper~\cite{nanoshaper} computed, respectively, the electrostatic potential surface and the solvent excluded surface (SES) to generate molecular surfaces, stored into VTK files. The APBS default parameters were used to generate surfaces, except for the point density (\textit{sdens}) at 1.00 due to the GMRES (Generalized Minimal Residual Method) limitation hard-coded by default.

\subsection{Performance evaluation}

To evaluate the performance of the methods on attributing the correct classes to the protein surface shapes, an in-house script has been developed to retrieve the predicted classes, perform the comparison with the ground truth, and compute classification metrics using Scikit-learn tools~\cite{scikit-learn}. To measure classification performance, several metrics were computed for each proposed method: accuracy and balanced accuracy scores, F1 score, Precision and Recall.

\textbf{Accuracy} classification score reflects the match between predicted labels and ground truth labels, here protein classes. The balanced accuracy can be used in case of imbalanced datasets.\\

\textbf{F1 score} (Eq \ref{eq:f1}), also known as F-measure, \textbf{Precision} (Eq \ref{eq:precision}) and \textbf{Recall} (Eq \ref{eq:recall}) correspond, respectively, to the following formulas: 

\begin{equation}
    F1 =\frac{2*TP}{2*TP+FP+FN},
    \label{eq:f1}
\end{equation}

\begin{equation}
    Precision = \frac{TP}{TP+FP},
    \label{eq:precision}
\end{equation}

\begin{equation}
    Recall = \frac{TP}{TP+FN},
    \label{eq:recall}
\end{equation}

where TP is the number of true positives, FP the number of false positives and FN the number of false negatives. An interpretation of the F1 score is the harmonic mean of Precision and Recall. \\

\section{Participating teams methods}\label{meth-ptcp}

The 15 methods from the 9 participating teams are summarized in Table \ref{tab:methods}. The details of the methods are presented in this section. Hardware and runtimes are presented in Supplementary Data.

\begin{table}[htbp]
\centering
\noindent
\begin{tabular}{|p{2.1cm}|p{2.8cm}|p{3.4cm}|p{3.5cm}|}
\hline
\textbf{Group} & \textbf{Features} & \textbf{Method Type} & \textbf{Previous SHREC tracks participation} \\
\hline
\authorN{Tatsuma}{Author\_1} & Geometry and Properties & Machine Learning & No \\
\hline
\authorN{Barisin \textit{et al.}}{Author\_2} & Geometry & Deep Learning & No \\
\hline
\authorN{Barisin \textit{et al.}}{Author\_2} & Geometry and Properties & Deep Learning & No \\
\hline
\authorN{Peng \textit{et al.}}{Author\_3} & Geometry & Deep Learning & No \\
\hline
\authorN{Kagaya \textit{et al.}}{Author\_4} & Geometry and Properties & Training free & 2019, 2021 \\
\hline
\authorN{Guerra \textit{et al.}}{Author\_5} (Methods 1 to 4) & Geometry & Deep Learning & No \\
\hline
\authorN{He \textit{et al.}}{Author\_6} (Methods 1 and 2) & Geometry & Deep Learning & No \\
\hline
\authorN{Li \textit{et al.}}{Author\_7} & Geometry & Deep Learning & No \\
\hline
\authorN{Tehrani \textit{et al.}}{Author\_8} & Geometry & Deep Learning & No \\
\hline
\authorN{Tehrani \textit{et al.}}{Author\_8} & Geometry & Machine Learning & No \\
\hline
\authorN{Yang \textit{et al.}}{Author\_9} & Geometry & Deep Learning & No \\
\hline
\end{tabular}
\caption{Overview of participating teams methods and used features (only geometry or geometry + electrostatic properties)}
\label{tab:methods}
\end{table}

\by{A. Tatsuma}{Author\_1}

\textbf{Statistics of local features and potentials.}

\github{https://github.com/yoshoku/shrec2025_protein}

To capture the complex shape features of protein models, we considered extracting local features and aggregating them into a feature vector. At first, the protein model was converted into a point cloud representation using Osada \textit{et al.} method~\cite{Osada2002}, which generates random points based on the vertices of the triangle mesh, and simultaneously, the potentials assigned to the vertices were also converted to random point potentials. Then, the Fast Point Feature Histogram (FPFH)~\cite{Rusu2009}, a local feature of the point cloud representation, was extracted. To aggregate local features into a feature vector for a protein model, statistics of the local features were calculated, such as mean vector and covariance matrix. The covariance matrix was vectorized by arranging its upper triangular elements. In addition, to incorporate protein-specific features, the histogram, mean, and variance of the potentials were calculated. The final feature vector was obtained by concatenating these features, followed by L2 normalization. In the experiments, the Sobol quasi-random sequence~\cite{Joe2003} was used for generating the point cloud representation consisting of 30,000 random points, and the number of bins of the potential histogram was set to 16.
For the classification phase, we used the Support Vector Machine (SVM) as implemented in Scikit-learn~\cite{Pedregosa2011}. In the experiments, the regularization parameter C was set to 32 and the radial basis function kernel was selected with a kernel coefficient gamma of 8. Moreover, since the dataset consists of imbalanced classes, the class\_weight parameter was set to “balanced”, which automatically adjusts the weights inversely proportional to the number of samples for each class. \\

\by{T. Barisin, E. Rusakov, U. Göbel}{Author\_2}

\textbf{Simplified RIConv++: Rotation Invariant Deep Network for Point Clouds. }

\github{https://github.com/ContactSoftwareAI/RINetwork-Shrec2025-Protein-Shape-Classification}

For this method, we used a simplified version of RIConv++ network~\cite{Zhang2022} with three simple rotation invariant features per layer. Original work extracts eight RI features on the nearest neighbors and relies on their “local” ordering. By keeping only three features, we removed the need to induce the order in the neighborhood and hence simplified feature extraction. Let $(p,n_{p})$ be a reference point with its normal and $(x,n_{x})$ a point in its local neighborhood. Then the following three rotation invariant features were extracted:
$$d\,=\,\left\| x-p \right\|,\,\,\alpha_{1}=\angle (\overline{xp},n_{p}),\,\,\alpha_{2}=\angle (\overline{xp},n_{x})$$

\begin{figure}
    \centering
    \includegraphics[width=0.98\linewidth]{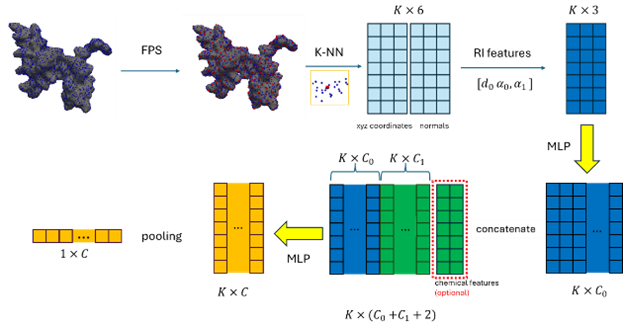}
    \caption{Steps of Rotation Invariant layer (RI Conv Layer)}
    \label{fig:Barisin_fig1}
\end{figure}

A Rotation Invariant layer (RI Conv Layer) consists of several steps, as shown in Figure \ref{fig:Barisin_fig1}.

From the present method, two architectures were tested: Run 1 used only geometric information (\authorN{Barisin\_v1}{Author\_2\_v1}), while Run 2 combined geometric and chemical information (\authorN{Barisin\_v2}{Author\_2\_v2}).

\begin{figure}
    \centering
    \includegraphics[width=0.98\linewidth]{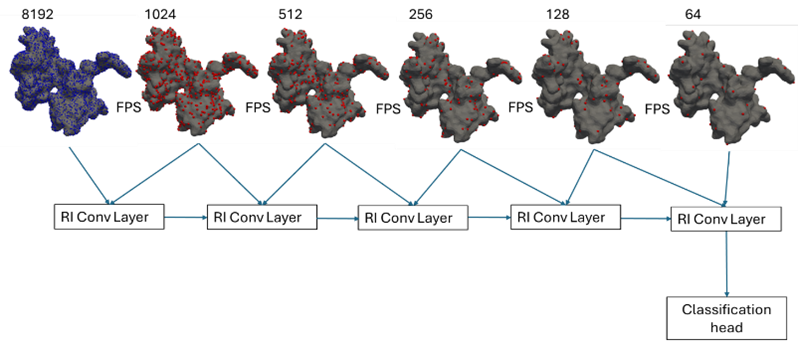}
    \caption{Network architecture of the simplified RIConv++ method}
    \label{fig:Barisin_fig2}
\end{figure}

Besides, potential and normal potential are both scalars and are considered to be rotation invariant. For Run 2, they were incorporated into the layer, enabling an efficient combination of geometric and chemical information. For network architecture, 5 RI layers were stacked to incorporate several scales of point clouds sampled by FPS algorithm (Figure \ref{fig:Barisin_fig2}). In total, network has 5.7 million parameters. \\

The training set was split into an 80/20 proportion and the last 20\% of data was used for validation.

\by{Y. Peng, S. Deng}{Author\_3}

\textbf{3D protein recognition based on 2D images.}

\github{https://github.com/thylove-dsq/shrec2025}

Starting from the three-dimensional structural model of proteins, we first converted the original VTK format into an OBJ format for subsequent image rendering processing. With the help of a graphics rendering engine, PyTorch 3D, protein surfaces were visualized and projected from multiple perspectives (here we take 8 directions because taking 8 images around the circle with equal curvature can obtain protein image features) to generate high-quality two-dimensional images, thereby achieving effective presentation of three-dimensional spatial information. It rotates around a circle with a 20-degree elevation angle, and is not evenly distributed on the sphere. It is not necessary to evenly distribute on the sphere because protein image features can be obtained by circling around the circle. Each OBJ file was rendered as 8 image samples from different perspectives. In theory, as long as the captured protein images are comprehensive enough, the features of the proteins can be fully extracted, so as to correctly retrieve proteins.

To alleviate the problem of model performance bias caused by class imbalance, an image enhancement strategy was adopted for classes with a small sample size, appropriately increasing the number of generated images to improve their representation ability and classification weight during the overall training process.

In the image feature extraction stage, we used GoogLeNet~\cite{Szegedy2015} deep convolutional neural network as the backbone model, calculated and added category weights (assign different weights to samples of different categories to balance the uneven distribution of samples in the dataset), and inputed the above two-dimensional images into the network for semantic feature extraction. Specifically, the output of the penultimate layer of GoogLeNet was extracted to obtain a high-level feature representation with a dimension of 1,024, which was used to characterize the spatial structural semantic information of proteins.

We used extracted image embedding vectors for K-nearest neighbor search, then calculated the nearest neighbor matching results of the test image in the training set feature space, and finally evaluated the classification accuracy of the structural representation based on this, thus verifying the effectiveness of rendering images and depth features in protein structure recognition tasks. \\

\by{Y. Kagaya, J.H. Park, D. Kihara}{Author\_4}

\textbf{3D-SURFER: Protein surface classification using 3D Zernike descriptors (3DZD).}

\github{https://github.com/kiharalab/SHREC2025}

The 3D Zernike descriptor (3DZD) is a representation that numerically represents 3D surface shapes as rotation-invariant feature vectors. We have a long history of applying 3DZD to a variety of tasks involving protein surface shapes and analyses of protein shapes~\cite{Han2019}. For example, 3D-SURFER~\cite{Aderinwale2022}~\cite{La2009}~\cite{Sael2008fast}~\cite{Han2017}, a tool that represents protein surface shapes using 3DZD to enable real-time retrieval of proteins with similar shapes, has been developed and is still maintained. This is also the case with another tool, EM-SURFER~\cite{Esquivel2015navigating}, which performs comparative searches of cryo-EM density maps using a similar approach. Furthermore, the LZerD docking protocol series~\cite{Venkatraman2009protein}~\cite{Esquivel2012multi}~\cite{Christoffer2021lzerd} was developed to predict protein-protein docking by comparing the 3DZD of two surfaces to identify shape complementarity. In addition, PL-PatchSurfer~\cite{Hu2014pl}~\cite{Shin2016pl}~\cite{Shin2024pl}, PatchSurfer~\cite{Sael2012detecting}~\cite{Sael2012constructing}, and PocketSurfer~\cite{Chikhi2010real} compare protein ligand-binding pockets and identify small molecules that can dock into them by assessing shape complementarity through 3DZD comparison. Based on this extensive expertise, we designed a shape retrieval and classification pipeline for the SHREC 2025 Protein Shape Retrieval track, building on the framework of 3D-SURFER.

The process of 3D-SURFER begins with a PDB file, and the protein surface is calculated using EDTsurf~\cite{Xu2009generating}. However, as we were provided with the 3D surface mesh information calculated by NanoShaper~\cite{Decherchi2013general} in VTK format, our initial step involved converting these files into PLY format, the native output format of EDTsurf. This conversion was performed using the Python VTK library. Subsequently, we followed the same steps as in 3D-SURFER to obtain the 3DZDs from these PLY surfaces. During this process, the surface mesh was converted into a binary voxel representation with a grid resolution of 1 Å from which the 3DZDs were computed. Given that electrostatic potential data was also provided, this information was utilized as well. From the provided potential, triangular mesh faces where all three vertices had a positive value were extracted, defining the "positive surface". Similarly, the "negative surface" was extracted by identifying triangular mesh faces with all three vertices having a negative value. For each of these three surfaces (all, positive, and negative), the 3DZD was computed up to an order of 20, resulting in three vectors, each with 121 values. The order of 20 was chosen to follow our previous studies that employed 3DZD for protein surface comparison~\cite{Aderinwale2022}~\cite{La2009}~\cite{Sael2008fast}~\cite{Han2017}. These three vectors were then concatenated, so that each surface was represented by a single vector of 363 values. This approach builds upon our previous methodology~\cite{Sael2008rapid}. The distance between two vectors was measured using the Euclidean distance of these vectors. These distances were computed for all pairs between the training and test datasets. For each test target, the label of the single training target with the smallest 3DZD distance was used as the predicted label. Notably, pairs of training and target samples with a volume difference exceeding 20\% were excluded from being used for label inference. This exclusion criterion was consistent with the approach used in EM-SURFER. \\

\by{M. Guerra, G. Palmieri, A. Ranieri, U. Fugacci, S. Biasotti}{Author\_5}

\textbf{Topological descriptors (\authorN{Guerra\_v1}{Author\_5\_v1}).}

\github{https://github.com/marcoguerra192/ProteinClassifier2025/}

In this method, each mesh was considered as a topological object in $\mathbb{R}^3$, either as a point cloud or as a triangulated surface. \\

\noindent\uline{Alpha descriptors.} In the literature, the Alpha complex~\cite{alpha-shape} filtration is widely used for biomolecule analysis, thanks to its relationship to the Voronoi diagram and its recognized ability to characterize the shape of the molecular surface~\cite{liang1998analytical}. Intuitively, the Alpha complex of an arbitrary point cloud is the simplicial complex consisting of points, edges, and triangles corresponding to non-empty intersections determined by simulating for each point of the cloud the presence of a ball of a certain radius depending on a parameter Alpha. We computed the Alpha filtration~\cite{alpha-shape} of the point cloud in $\mathbb{R}^3$, and its persistent homology in dimension $k=0,1,2$. The resulting $k$-persistence diagrams were then thresholded at longer lifespans, and for each $k$ a 5x5 persistence image~\cite{adams2017PersImages} was used for vectorization, resulting in a feature vector of length 75. \\

\noindent\uline{Radial descriptors.} The centroid of the point cloud was computed. A reference sphere was subdivided into 8 equal spherical sectors. We considered the intersection of the surface with each sector. For each, we computed the [.25, .5, .75] quantiles of the distribution of distance of points from the centroid, the cumulative radial potential at those quantiles, the number of persistent homology classes whose lifespan was at least one tenth of the median, the birth and death time of the longest finite $H_0$ bar, and the potentials of the generating vertices in the lower-star filtration. This resulted in 96 features. \\

\noindent\uline{Data preparation.} The descriptors were combined for each protein. Rotational invariance of the classifier was enforced by augmenting each data point by the action of the group of rotations onto the radial descriptors. Given our subdivision into 8 sectors, 8 different descriptions were obtained for each protein. The descriptors were de-correlated and scaled. The strong imbalance between the largest and smallest classes is troublesome for classification purposes. The dataset was partially balanced by introducing duplicates of the more infrequent classes. Given the size difference between the largest and smallest classes, a perfect balancing would have required an unreasonable amount of duplicates. Finally, the data was divided into 80/20 training and validation subsets. \\

\noindent\uline{Classification.} We used a small neural network with 4 fully-connected layers, whose sizes decreased linearly from the input to output size, with ReLU activations. Each layer employed dropout with $p=.25$, and we further enforced $L^2$-regularization. \\

By using data augmentation on the test set as well, a set of 8 predictions for each protein to classify was obtained. The result was then given by majority, with ties broken by a priori class frequency. \\

\textbf{PointNet architecture (\authorN{Guerra\_v2}{Author\_5\_v2}).}

\github{https://gitlab.com/ml845216/Protein-Classification}

\noindent\uline{Dataset Construction and Preprocessing.} For dataset construction, we utilized protein point clouds subjected to minimal data augmentation. Applied transformations included random spatial translation and additive Gaussian noise (empirical observations indicated that protein rotation adversely impacted performance). All point clouds were normalized to fit within a unit sphere and processed using a random sampling strategy to ensure uniform point density. \\

\noindent\uline{Model Architecture and Training Protocol.} A standard PointNet~\cite{qi2017pointnetdeeplearningpoint} architecture (26M parameters) initialized from scratch was employed. Pre-training experiments revealed no significant improvement in training or validation loss trajectories, with final accuracy remaining comparable to models trained without pre-training. \\

To mitigate class imbalance, a class-balancing strategy was applied: samples from under-represented classes were duplicated by a factor proportional to the ratio of the largest class size, capped at a maximum multiplier of 10. This approach preserved dataset diversity while reducing bias toward dominant classes.

To obtain the final prediction, we employed three PointNet models at three distinct stages of the training process: at epoch 80, at the conclusion of the training phase, and at the point of minimal validation loss. By integrating the three distinct logit vectors generated at these stages and aggregating them via summation, the ultimate prediction was derived. \\

\textbf{A multi-view image-based approach using a ViT encoder (\authorN{Guerra\_v3}{Author\_5\_v3}).}

\github{https://gitlab.com/ml845216/Protein-Classification/-/tree/master/ViT}

For this third submission, we developed a novel image-based pipeline for protein structure classification. Starting from VTK files, we implemented a PyVista-based script to generate 1,024×1,024-pixel RGB images. Each image comprised an 8×8 grid of 128×128-pixel sub-views, captured at 45-degree intervals along both azimuth and elevation angles to comprehensively represent the protein 3D structure. \\

\noindent\uline{Dataset Curation and Splitting.} We balanced the original dataset by first computing a target average per class from the total desired "budget" of 50,000 samples for the final balanced dataset. We then up-sampled smaller classes to this average via random selection with replacement, multiplying their occurrences up to 1,000 times. To preserve original dataset diversity, we did not down-sample over-represented classes. The final balanced dataset is further subjected to data augmentation. The dataset was partitioned into training and validation sets using an 80/20 random split to preserve class distribution. \\

\noindent\uline{Model Architecture and Training.} We trained a classification model using a \textit{vit-giant-patch14-reg4-dinov2} encoder~\cite{oquab2023dinov2}, implemented via a PyTorch and Fast.ai~\cite{jeremy_fastai} pipeline inside Jupyter notebook. To enhance generalization, a custom data augmentation strategy was applied:
\begin{itemize}
    \item \emph{Stochastic Sub-Image Rotation}: each 128×128 sub-view in the 8×8 grid was rotated by a random angle (0–359°), with the composite image reconstructed post-rotation.
    \item \emph{RGBShift Augmentation}: targeted shifts were applied to the red and green channels, reflecting their dominance in the PyVista-rendered protein visualizations.
    \item \emph{Resolution Scaling}: images were finally resized to 518×518 pixels to match the encoder patch dimensions.
\end{itemize}

This approach aims at demonstrating the efficacy of multi-view representation learning for 3D biological structures, hopefully achieving robust performance while addressing both dataset complexity and class imbalance challenges. \\

\textbf{Combining different approaches (\authorN{Guerra\_v4}{Author\_5\_v4}).}

\github{https://gitlab.com/ml845216/Protein-Classification}

As a fourth submission, we chose a combined classification obtained by a weighted majority vote among the three methods previously described. Specifically, for each protein, we had three proposed classes, each with a confidence score. In case a majority existed, that class was chosen. If no majority existed, our algorithm excluded the class with lower confidence and kept the class with a priori highest frequency (based on the training set).

\by{R. He, H. Benhabiles, A. Cabani, K. Hammoudi}{Author\_6}

\textbf{3D-PROSPER: 3D Protein Representation via Optimized Self-supervised and multi-task Point cloud Encoding for Recognition.}

\github{https://github.com/cabani/3D-PROSPER}

\begin{figure}
    \centering
    \includegraphics[width=0.98\linewidth]{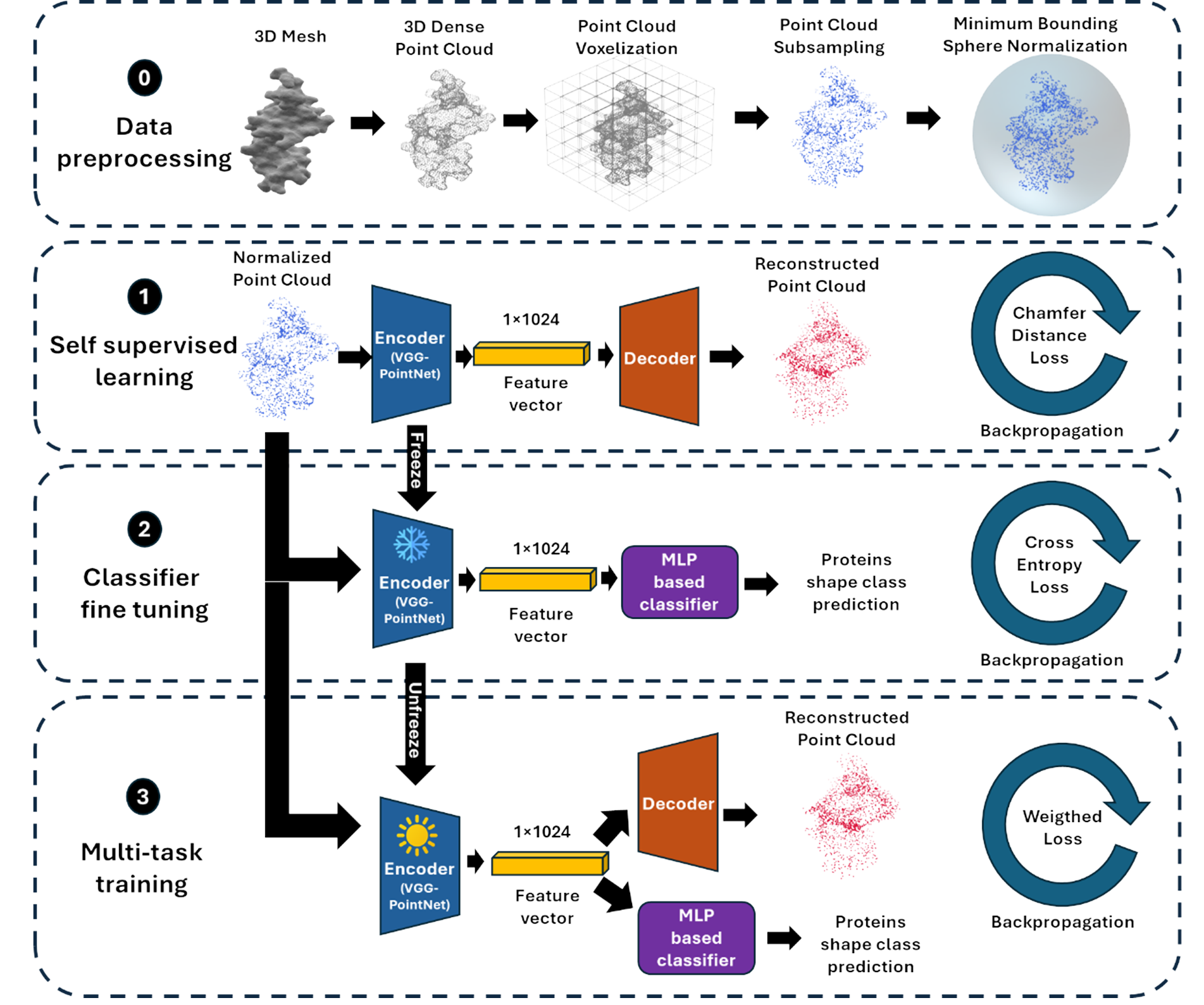}
    \caption{Overview of the 3D-PROSPER framework for protein shape classification}
    \label{fig:He_fig}
\end{figure}

Our framework consists of four key stages: data preprocessing, self-supervised learning, classifier fine-tuning, and multi-task training. Each step incrementally builds towards a robust and accurate protein shape predictor, as illustrated in Figure \ref{fig:He_fig}. \\

\noindent\uline{At step 0 (Data Preprocessing)}, the 3D mesh representation of each protein was converted into a point-cloud format. Specifically, a voxel-based subsampling method was applied to uniformly sample 2,048 points from the mesh~\cite{Benhabiles2013}. Voxel-based sub-sampling ensured geometric consistency and uniformity across protein structures, while sampling 2,048 points made it suitable for our deep learning backbone. The resulting point cloud was then normalized into a zero-centered unit sphere, using the minimum bounding sphere of each protein for consistent scaling and alignment. This step ensured geometric invariance in terms of scale and translation as well as standardization across the dataset. \\

\noindent\uline{At step 1 (Self-Supervised Learning)}, to learn meaningful representations from unlabeled data, we employed a self-supervised auto-encoder for protein point cloud reconstruction. The encoder, called VGG-ProteinNet, is inspired by the VGG architecture~\cite{Simonyan2014} in terms of depth and convolutional structure, and integrates T-Net modules from PointNet~\cite{Qi2017} to enhance geometric robustness at each convolutional layer. The decoder reconstructs the original point cloud from the learned latent space. The network was trained using the Chamfer Distance loss~\cite{Fan2016}, encouraging the reconstruction to faithfully preserve structural details of the original protein shape, regardless of its class. \\

\noindent\uline{At step 2 (Classifier Fine-Tuning)}, once the encoder had learned a robust latent representation, a Multi-Layer Perceptron (MLP) classifier was attached on top of it. During this phase, the encoder weights were frozen, and the MLP was trained using a cross-entropy loss to predict the protein shape class (3D-PROSPER\_v1). This step focuses on learning a discriminative mapping from the learned latent space to the target shape classes. \\

\noindent\uline{At step 3 (Multitask Training)}, to further enhance classification performance, we adopted a multitask learning strategy where both reconstruction and classification objectives were optimized jointly. In this phase, we unfreezed the encoder, and simultaneously fine-tuned the auto-encoder (reconstruction decoder) and the MLP classifier. The training was driven by a weighted sum of the Chamfer Distance loss and the cross-entropy loss, allowing the model to retain structural fidelity while improving its shape classification capability (3D-PROSPER\_v2). The entire framework was trained on the dataset provided by the track, applying an 80\%/20\% train-validation split with stratification to maintain the original class distribution. \\

\by{H. Li, H. Huang, C. Li}{Author\_7}

\textbf{RISurProtNet.}

\github{https://github.com/a1796241465/RISurProtNet/tree/The-first}

Our method is built upon the RISurConv framework proposed by Zhang \textit{et al.}~\cite{Zhang2024}, which enables efficient and rotation-invariant deep learning on 3D point clouds. The core idea of RISurConv is to represent local geometry using surface patches rather than treating the point cloud as an unordered set of points. \\

\noindent\uline{At step 1 (Surface patch construction)}, for each point in the point cloud, a local surface patch was constructed by connecting it with its neighboring points to form triangle meshes. This surface-based representation preserved fine-grained geometric details. \\

\noindent\uline{At step 2 (Feature extraction using RISP)}, from each local surface patch, a 14-dimensional vector of Rotation-Invariant Surface Properties (RISP) was computed. These features included relative angles, distances, and normal orientations between adjacent triangles. As they rely on local geometric relationships rather than absolute positions, the RISP descriptors are inherently invariant to arbitrary SO(3) rotations. \\

\noindent\uline{At step 3 (Feature embedding and refinement)}, the RISP vectors were embedded into a high-dimensional space using shared multilayer perceptrons (MLPs). To further capture contextual information, self-attention modules were applied to refine the embedded features and model interactions between neighboring surfaces. \\

\noindent\uline{At step 4 (Hierarchical encoding)}, the network stacked five RISurConv layers to extract hierarchical features. Each layer contained Farthest Point Sampling (FPS) to select representative anchor points, K-Nearest Neighbors (KNN) to define local surface patches, two Self-Attention layers to enable context-aware feature refinement, feature concatenation and transformation via shared MLPs, and max pooling to aggregate local surface information. \\

\noindent\uline{At step 5 (Global representation and classification)}, the hierarchical features were aggregated and passed into a Transformer encoder~\cite{Vaswani2017} to model long-range dependencies across the surface. The output was processed by fully connected layers to predict the class label for the protein structure. This architecture is naturally invariant to both rotation and point permutation, without requiring extensive data augmentation. Compared to other rotation-invariant methods such as RIConv~\cite{Zhang2019}, RI-GCN~\cite{Cai2022}, and RIConv++~\cite{Zhang2022},
RISurConv achieves superior classification accuracy and improved robustness.
To better align with the SHREC 2025 protein dataset characteristics, the following modifications were applied:
\begin{itemize}
    \item \emph{Data Uploading Pipeline Modification}: the data loading process was customized to support protein-specific 3D point clouds derived from VTK files. The pipeline parsed atomic coordinate data, applied preprocessing steps such as coordinate normalization and farthest point sampling (FPS), and enabled flexible configuration of input formats. This setup ensured structural compatibility with the RISurConv framework while accommodating the unique characteristics of biomolecular data.
    \item \emph{Feature Extension}: the RISurConv input was extended by incorporating auxiliary features such as potential and normal potential, in addition to the spatial coordinates. These features were concatenated with RISurConv extracted feature representation and fed through the original network pipeline.
\end{itemize}

The method leverages the inherent rotation-invariance of RISurConv for robust feature extraction from molecular geometry, while our custom extensions address domain-specific challenges of protein data. \\

\by{A. Tehrani, F. Meng, F. Heidar-Zadeh}{Author\_8}

\textbf{Siamese Deep Metric Learning on Property-Spectrum (\authorN{Tehrani\_v1}{Author\_8\_v1}).}

\github{https://github.com/qtchem/shrec2025-ml-protein-surfaces}

Due to the highly imbalanced nature of the dataset, our surface descriptors were first transformed into a learned embedding via deep metric learning, then these embeddings were classified with a feed-forward neural network. Deep metric learning involved learning an embedding space where protein surfaces that are of the same class were transformed to be closer together, and otherwise were spread more apart. This was achieved from utilizing a Siamese neural network built from residual fully connected blocks, composed of Linear, Batch-Norm, ReLU, and dropout layers. To train the Siamese network, a triplet loss function was used, where a random anchor sample was selected, then a positive example (same class) and a negative example (different class) were randomly selected, ensuring that the anchor was closer to the positive example and further apart from the negative example.

The Siamese network was trained for 50 epochs, then its weights were freezed and a feed-forward classifier was trained on the learned embeddings using cross-entropy loss. The network was composed of repeating Linear, Batch-Norm, ReLu, and dropout layers, respectively. After a set number of classifier epochs, we resumed Siamese training to learn a better embedding space. We then applied hard-example mining by increasing the sampling probability of incorrectly classified protein surfaces, so that the triplet loss focused on “difficult” examples.

Our initial input features used the property-spectrum approach to represent the protein surface, and its surface property function (e.g. its potential) as a spectrum-like (wave) descriptor. This is a novel, alignment-free, property-dependent shape descriptor based on the Laplace-Beltrami operator. Our method is soon to be submitted for publication and to be posted on arxiv. We consider the following list of surface-properties to produce various kinds of property-spectra: electrostatic potential, Gaussian curvature, principal curvature, shape index, radial distance to its center, and normal direction in x,y,z direction. We stratified the data into 90/10 validation sets, extracted 40 PCA components per spectrum, concatenated them and applied Min–Max scaling. \\

\textbf{Gradient Boosting LightGBM Model on Property-Spectrum (\authorN{Tehrani\_v2}{Author\_8\_v2}).}

\github{https://github.com/qtchem/shrec2025-ml-protein-surfaces}

For the second method, we also used the light gradient-boosting machine (LightGBM) model with the same property-spectra features generated above. The exact same procedures were used, including Min-Max scaling and principal component analysis for the same selected spectrum features. Data was stratified into 90/10 training and test sets. \\

\by{T.A. Yang}{Author\_9}

\textbf{ProtoCluster.}

\github{https://github.com/YangTuanAnh/ProtoCluster}

\begin{figure}
    \centering
    \includegraphics[width=0.9\linewidth]{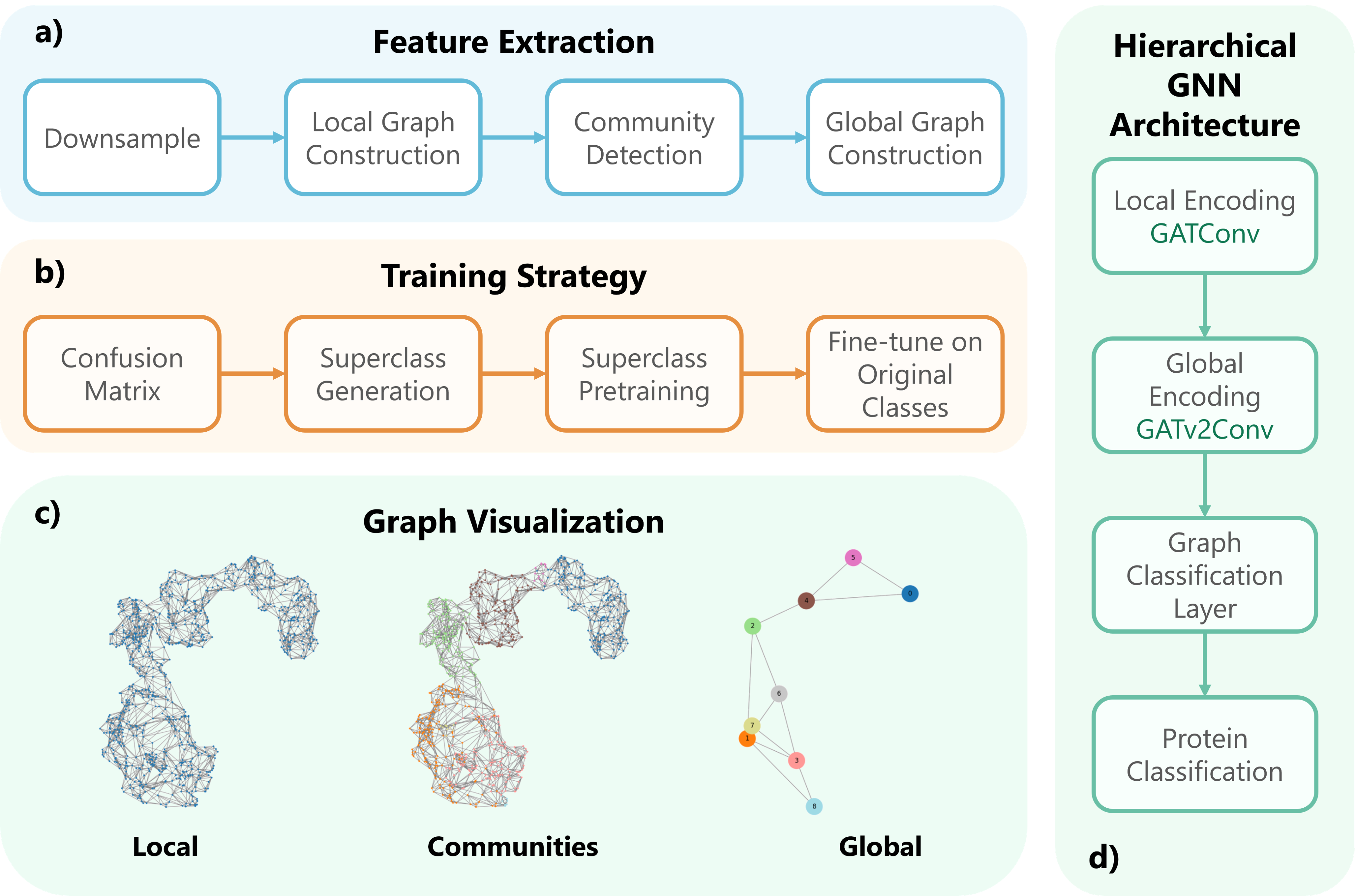}
    \caption{Overview of the ProtoCluster pipeline}
    \label{fig:Yang}
\end{figure}

In this method, we constructed a hierarchical graph representation from 3D geometric meshes stored in VTK files, with each mesh representing a single protein. The pipeline consisted of three stages (Figure \ref{fig:Yang}): local graph construction from mesh vertices, community detection for subgraph decomposition, and global graph construction from inter-community relationships. Implementation was performed using PyVista, NetworkX, and PyTorch Geometric. Meshes were loaded using pyvista.read() and downsampled to 1\% of the original vertex count via decimate\_pro() (reduction factor: 0.99) to reduce computational cost while preserving topology. The downsampled mesh was converted into an undirected local graph by assigning each vertex to a node and connecting vertices that form triangular mesh faces. Nodes were annotated with the following features: position (pos): 3D coordinates, normal (normals): surface normal vectors, potential (potential): scalar field (e.g.: electrostatic potential), normal potential (normal\_potential): modulated scalar based on geometry, encoded position (encoded\_pos): sinusoidal encoding of coordinates, in total 14 features.

To capture local structure, the greedy modularity maximization algorithm was applied to partition the graph into communities. Each community formed a sub-graph representing a spatially localized node cluster. A global (cluster) graph was constructed by representing each community as a node. The spatial position of each node was defined as the center of mass of its vertices. Edges were added between communities that were connected in the original graph, with edge weights given by the Euclidean distance between community centers. Each global node was further annotated with aggregated attributes: mean potential, mean normal potential, average normal vector, and encoded center position.

We design a graph-based classification architecture that captures both local patch-level representations and global community-level structure. At the local level, each patch was modeled as a node in a graph, and two layers of GATConv were used to learn context-aware patch embeddings. These local features were then propagated into a higher-level community graph, where two layers of GATv2Conv and a classifier head were applied to perform the final graph-level classification. All graph convolutional layers used a fixed hidden dimension of 128. To improve generalization, we employed a superclass-based pre-training strategy: coarse class labels were obtained via hierarchical clustering of the confusion matrix, enabling the model to pre-train on superclass labels before fine-tuning on the original fine-grained labels. \\

\section{Results}\label{res}

In the following sections, for the sake of clarity, we used the name of the proposed method if provided or the name of the first author of the participating team. \\

\begin{figure}[htbp]
    \centering
    \includegraphics[width=0.9\linewidth]{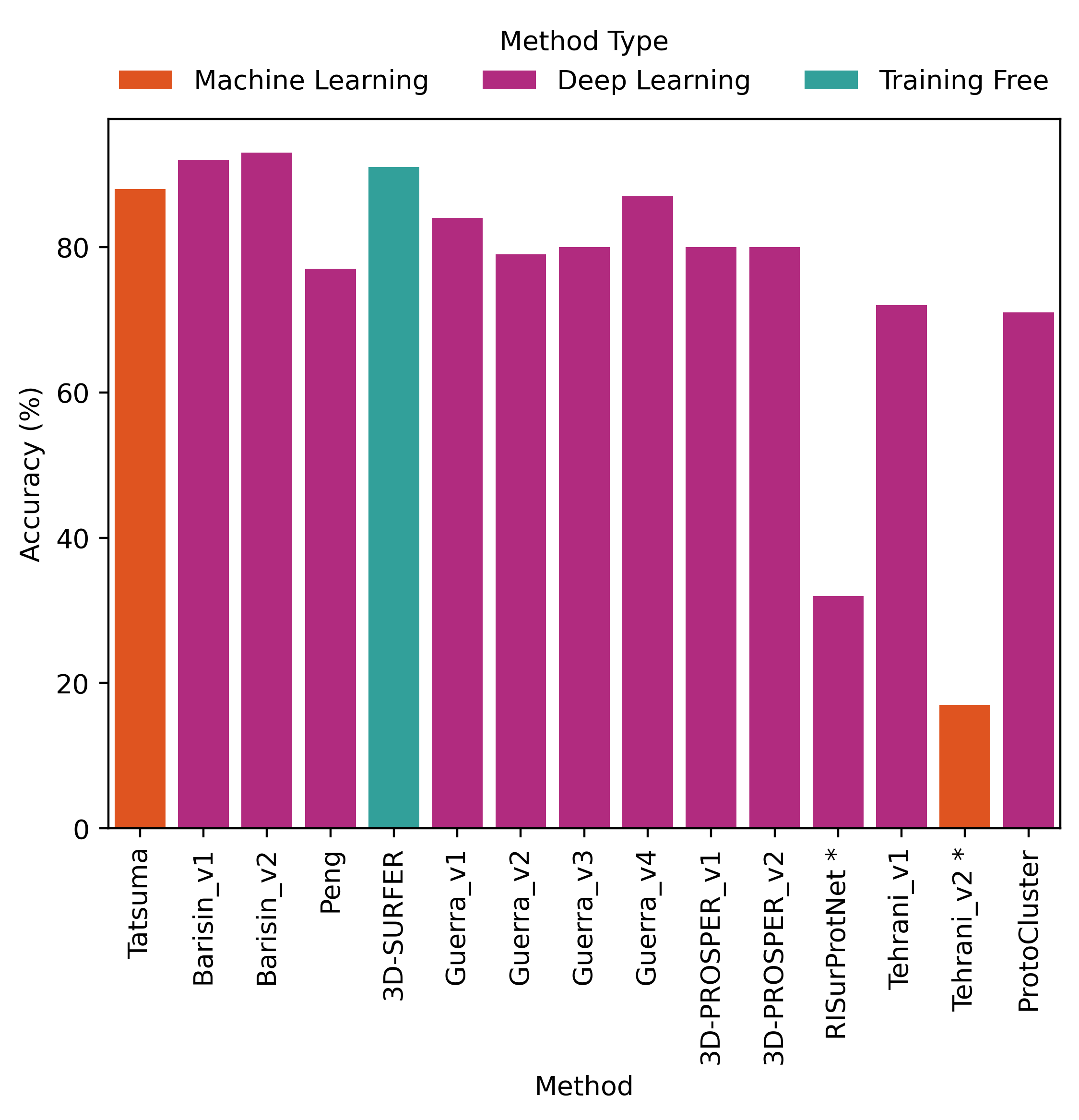}
    \caption{Accuracy percentage computed for each method. *: Methods with very low scores, due to technical problems that will be explained in Discussion}
    \label{fig:Accuracy}
\end{figure}

\begin{figure}[htbp]
    \centering
    \includegraphics[width=0.9\linewidth]{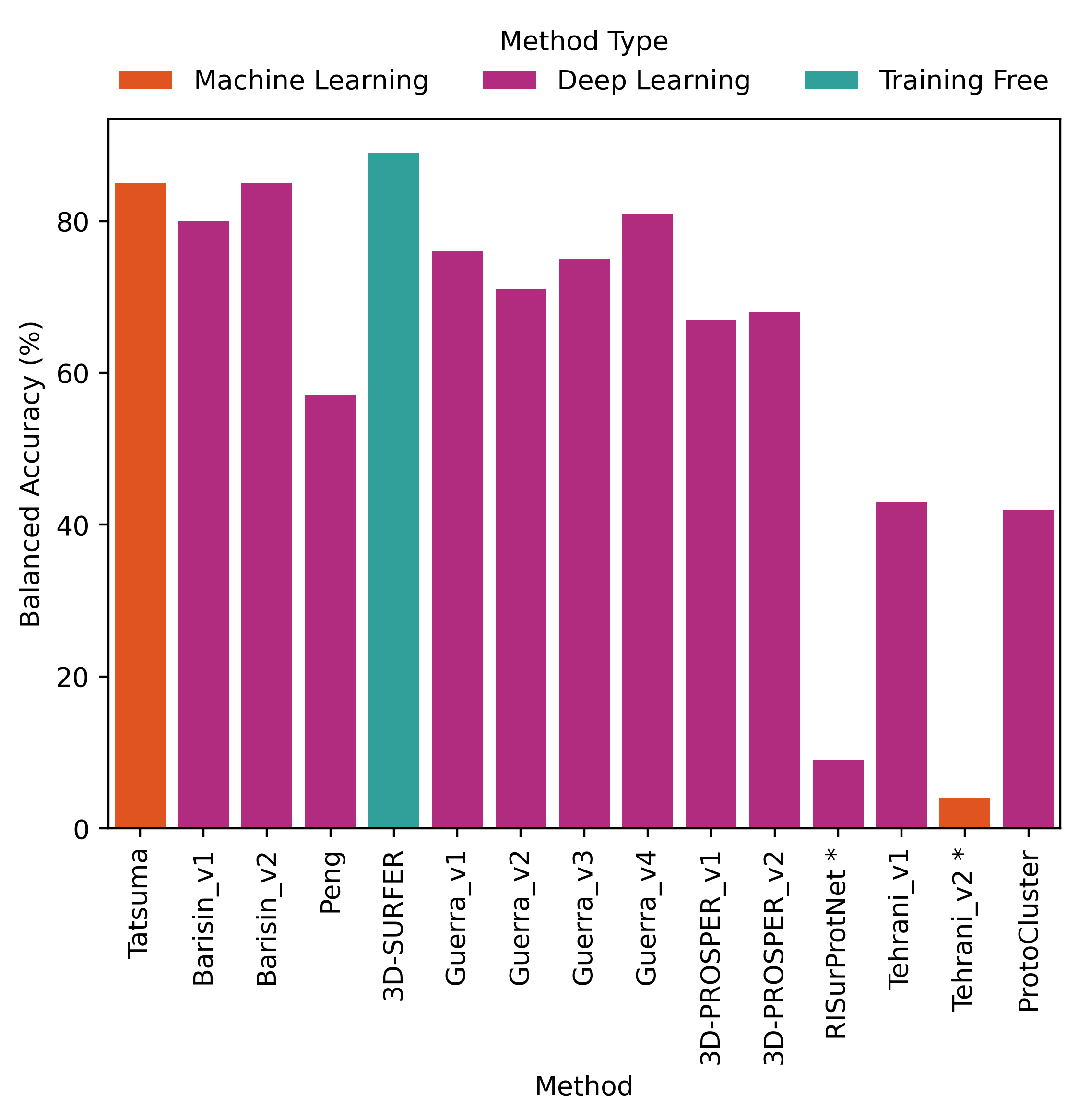}
    \caption{Balanced accuracy percentage computed for each method. *: Methods with very low scores, due to technical problems that will be explained in Discussion}
    \label{fig:Balanced_accuracy}
\end{figure}

\begin{figure}[htbp]
    \centering
    \includegraphics[width=0.9\linewidth]{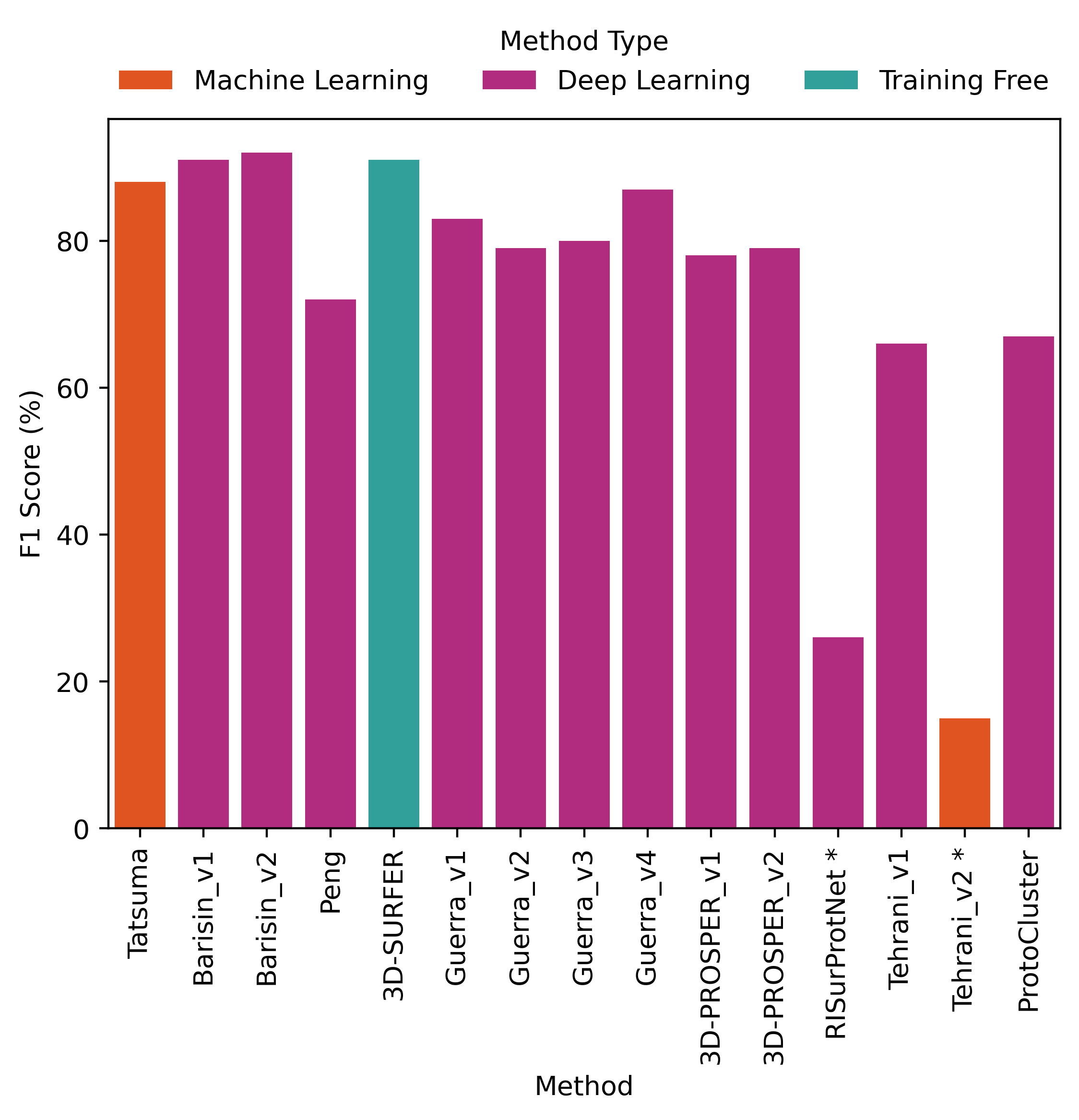}
    \caption{F1 score percentage computed for each method. *: Methods with very low scores, due to technical problems that will be explained in Discussion}
    \label{fig:F1_score}
\end{figure}

\begin{figure}[htbp]
    \centering
    \includegraphics[width=0.9\linewidth]{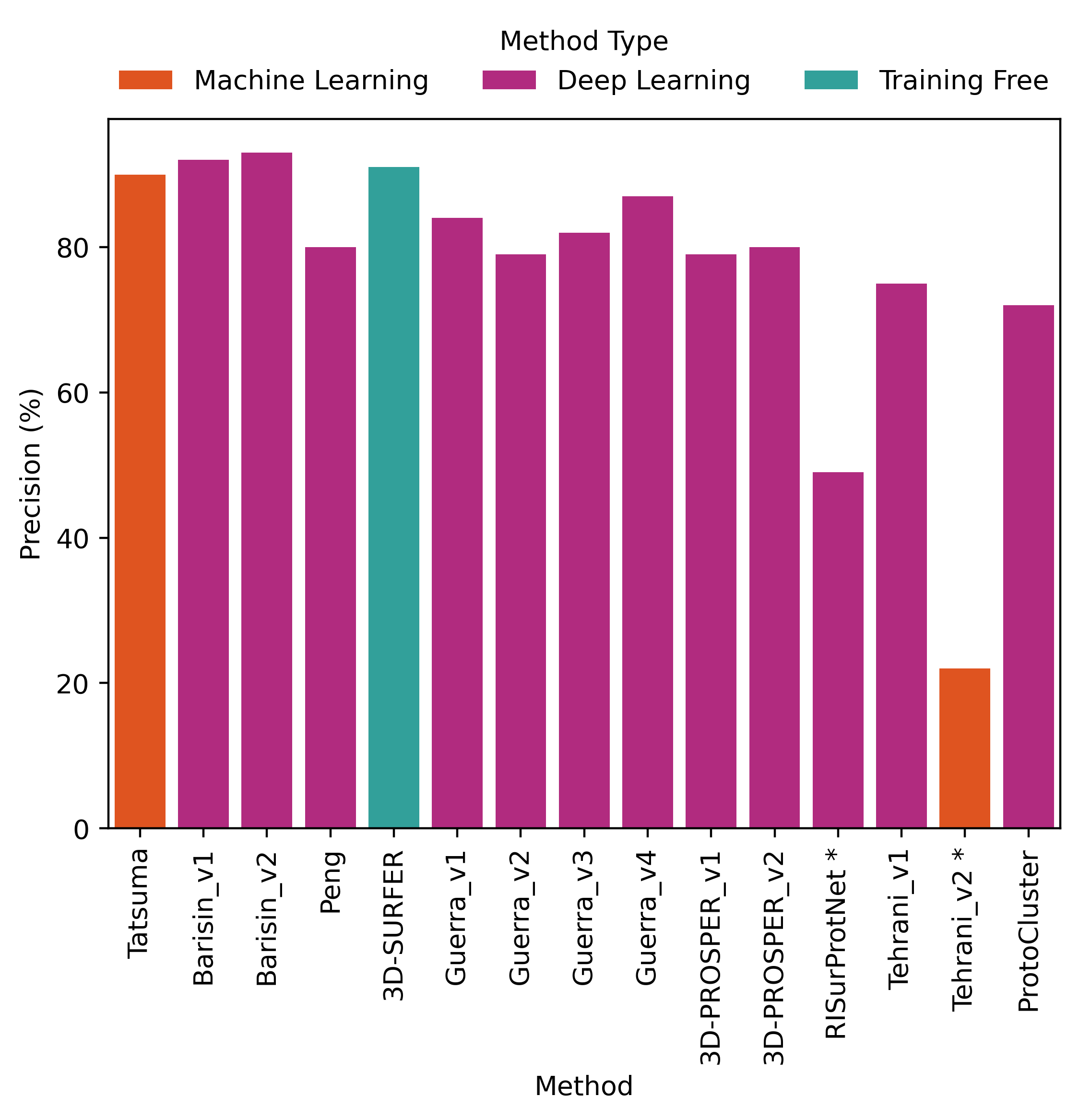}
    \caption{Precision percentage computed for each method. *: Methods with very low scores, due to technical problems that will be explained in Discussion}
    \label{fig:Precision}
\end{figure}

\begin{figure}[htbp]
    \centering
    \includegraphics[width=0.9\linewidth]{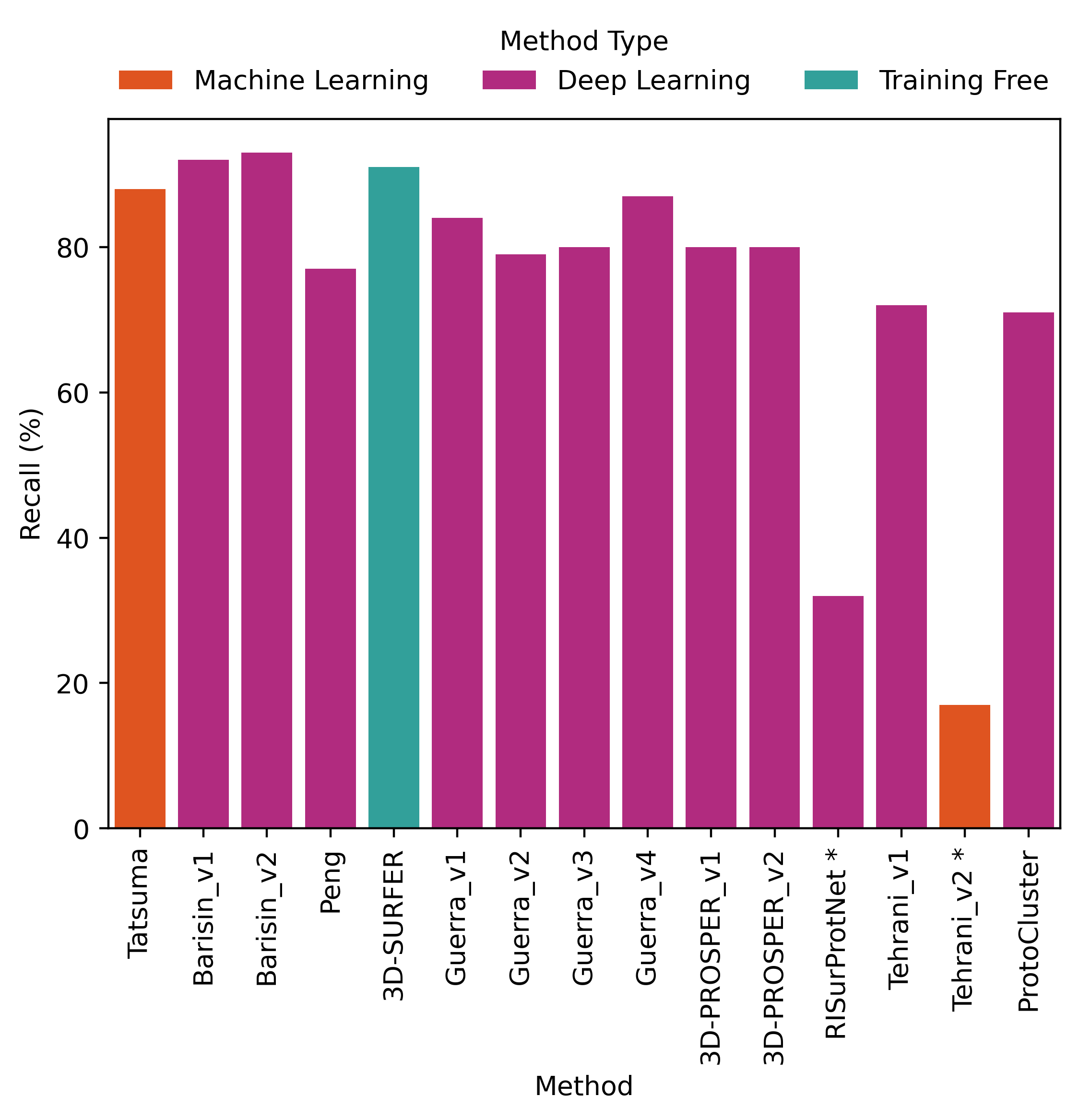}
    \caption{Recall percentage computed for each method. *: Methods with very low scores, due to technical problems that will be explained in Discussion}
    \label{fig:Recall}
\end{figure}

The comparative evaluation of the performance in retrieval of the different proposed methods was achieved by computing the metrics presented in Section 2. The values are presented as histograms for accuracy (Figure \ref{fig:Accuracy}), balanced accuracy (Figure \ref{fig:Balanced_accuracy}), F1 score (Figure \ref{fig:F1_score}), Precision (Figure \ref{fig:Precision}) and Recall (Figure \ref{fig:Recall}).


Most proposed methods (all but 2) displayed an accuracy (Figure \ref{fig:Accuracy}) ranging from 71\% to 93\% \textit{i.e} correct matches between predicted class and ground truth class. The two remaining proposed methods displayed an accuracy of 32\% and 17\%. This reflects that the proposed methods (except 2) correctly classified the protein surfaces and identified the correct classes with success.

All proposed methods displayed similar values for F1 score (Figure \ref{fig:F1_score}), Precision (Figure \ref{fig:Precision}) and Recall (Figure \ref{fig:Recall}). F1 score values ranged from 66\% to 92\% (except for RISurProtNet with 26\%  and for \authorN{Tehrani\_v2}{Author\_8\_v2} method with 15\%), Precision ranged from 72\% to 93\% (except for RISurProtNet with 49\% and for \authorN{Tehrani\_v2}{Author\_8\_v2} method with 22\% ). Recall ranged from 71\% to 93\% (except for RISurProtNet with 32\%  and for \authorN{Tehrani\_v2}{Author\_8\_v2} method with 17\%). All methods but two were able to retrieve the majority of the correct protein classes. Overall, \authorN{Barisin\_v1}{Author\_2\_v1} and \authorN{Barisin\_v2}{Author\_2\_v2} displayed the best performance in retrieval.

For to the balanced accuracy (Figure \ref{fig:Balanced_accuracy}), which takes into account the imbalance of the dataset between classes, different trends were observed. As expected, compared to accuracy, all proposed methods displayed a lower balanced accuracy. Two of the proposed methods (3D-SURFER and \authorN{Tatsuma}{Author\_1} method) respectively displayed a balanced accuracy with 2\% and 3\% reduction compared to accuracy, outperforming or equalizing the best proposed methods. All other proposed methods maintained their respective rankings in performance according to balanced accuracy.\\

The F1 score per class was analyzed for each proposed method. A stacked-F1 score between 0.0 and 15.0 (Supplementary Data: Figure 2) was computed for each class. A stacked-F1 score of 15.0 indicates that the proposed methods achieved an F1 score of 1. All proposed methods displayed a low F1 for class 36 (stacked-F1 score of 2.47), class 73 (stacked-F1 score of 1.0), class 78  (stacked-F1 score of 0.4), and class 80  (stacked-F1 score of 1.67). All proposed methods achieved an F1-score less than 50\% for classes 78 and 80, 14 proposed methods for class 73, and 12 proposed methods for class 36. \authorN{Guerra\_v1}{Author\_5\_v1} is the only proposed method that achieved 100\% in Recall for class 78, despite the F1 score and Precision of about 0\%.

Class 8 was the largest class with 514 protein surfaces in the test set. All proposed methods displayed a F1 score between 90\% and 100\% (except for \authorN{Tehrani\_v2}{Author\_8\_v2} method with 40\% and RISurProtNet with 70\%). Four proposed methods (3D-PROSPER, \authorN{Peng}{Author\_3} method, \authorN{Guerra\_v1}{Author\_5\_v1} method and \authorN{Guerra\_v4}{Author\_5\_v4} method) displayed a Recall of 100\% despite F1 score below 100\%. F1 score of 100\% was observed for \authorN{Barisin\_v1}{Author\_2\_v1} and \authorN{Barisin\_v2}{Author\_2\_v2} methods, 3D-SURFER and \authorN{Tatsuma}{Author\_1} method. \\

All proposed methods achieved Recall below 100\% for 25 out of 97 classes (Supplementary Data: Figure 3). The proteins 6epc\_14:N:7 (class 29) and 6x5a\_4:H:H (class 34) were not correctly predicted by all proposed methods in comparison with ground truth classes. \\

The test set included nine empty classes (26, 42, 44, 50, 58, 63, 72, 77 and 95). They included 2 objects in the train set each. \authorN{Barisin\_v2}{Author\_2\_v2} method, 3D-SURFER, 3D-PROSPER\_v1, \authorN{Guerra\_v1}{Author\_5\_v1} method, \authorN{Guerra\_v4}{Author\_5\_v4} method, RISurProtNet and \authorN{Tehrani\_v1}{Author\_8\_v1} method were the ones which successfully assigned no element to these classes. The eight other proposed methods predicted between 1 and 2 elements in these classes.

\section{Discussion}\label{disc}

\noindent\textbf{Performance overview.} The participating teams proposed promising methods applied to the field of protein shape classification.

Three proposed methods displayed excellent results, with more than 90\% of accuracy and F1 score. Two of them used a deep learning algorithm (\authorN{Barisin\_v1}{Author\_2\_v1} and \authorN{Barisin\_v2}{Author\_2\_v2} methods, using point cloud data representation combined with rotation invariant deep neural networks), the other used a training-free approach (3D-SURFER). Six proposed methods reached high performance scores, with more than 80\% of accuracy and F1 score: \authorN{Tatsuma}{Author\_1}, \authorN{Guerra\_v1}{Author\_5\_v1}, \authorN{Guerra\_v3}{Author\_5\_v3} and \authorN{Guerra\_v4}{Author\_5\_v4} approaches, and both 3D-PROSPER versions. \authorN{Peng}{Author\_3}, \authorN{Guerra\_v2}{Author\_5\_v2}, \authorN{Tehrani\_v1}{Author\_8\_v1} and ProtoCluster obtained very good results by using deep learning algorithms, getting more than 70\% of accuracy and F1 score. \\

Two approaches (RISurProtNet and \authorN{Tehrani\_v2}{Author\_8\_v2}) displayed low performance with accuracy and F1 score below 50\%. For RISurProtNet, the result was explained by hardware limitations and time constraints. The training was insufficient with 55 out of 450 epochs only. But we consider that an accuracy of 32\% with only 55 epochs was a decent result. The second explanation was that the under-represented classes were not taken into account. Results submitted after the deadline displayed a significant improvement in balanced accuracy (26\% according to the new results against 9\% initially, see Table 1 in Supplementary Data) with the introduction of a class-weighted module.

For \authorN{Tehrani\_v2}{Author\_8\_v2} method, the test set was not processed properly. Standardization and principal component analysis (PCA) were performed for the different types of features they calculated. But when running the predictions, the correct transformations for each feature were not used. After correction, the new accuracy was significantly better (61\% according to the new results against 17\% initially, see Table 2 in Supplementary Data). 

Although further investigation will be necessary, these explanations were important points to consider (take into account the dataset imbalance, avoiding underfitting) for future developments, and therefore made an important contribution. \\

Two participating teams proposed several methods for this challenge. \authorN{Guerra \textit{et al.}}{Author\_5} team proposed 4 methods (\authorN{Guerra\_v1}{Author\_5\_v1}, \authorN{Guerra\_v2}{Author\_5\_v2}, \authorN{Guerra\_v3}{Author\_5\_v3}, \authorN{Guerra\_v4}{Author\_5\_v4}). The fourth was a combination of the classification of the first three. This combined classification reached the highest scores for all classification metrics among these 4 methods. Accuracy and balanced accuracy reached 87\% and 81\% respectively, and F1 score, Precision and Recall increased to 87\%, \textit{i.e.} a performance gain of 3\% to 10\%. The significant improvement can thus be achieved by combining different approaches.

The \authorN{He \textit{et al.}}{Author\_6} team also proposed two versions of the 3D-PROSPER method. The difference of performance between both of them was very small. The accuracy was maintained at 80\%, but the second proposed version improved of 1\% the balanced accuracy, F1 score and Precision. Both implementations thus reached a high-level performance of classification. \\

In comparison to ML/DL methods, the training-free method 3D-SURFER performed very well. 3D-SURFER was situated in the best rankings of performance. Accuracy was equal to 91\%, a difference of 1\% and 2\% compared to \authorN{Barisin\_v1}{Author\_2\_v1} and \authorN{Barisin\_v2}{Author\_2\_v2} methods, respectively. 3D-SURFER kept high results for balanced accuracy, as well as \authorN{Tatsuma}{Author\_1} method. The dataset imbalance was therefore well managed by these proposed methods. \\

\noindent\textbf{Class-wise performance.} An F1-score below 50\% is considered as not good performance. All proposed methods displayed an F1-scores below 50\% for classes 78 and 80. The stacked-F1 scores were 40\% and 17\%, respectively.

12, 13 and 14 proposed methods displayed an F1 score of less than 50\% for classes 34, 36 and 73 respectively. For class 78, F1 score was near 0 for all methods. Recall was equal to 100\% for \authorN{Guerra\_v1}{Author\_5\_v1} method only, \textit{i.e.} all true positives were found. Precision was close to 0 and caused a drop in F1 score for all the proposed methods. It was globally the same observation for class 73. Only 3D-SURFER performed with an F1 score equal to 100\% for this class. For class 80, all proposed methods failed in Precision and Recall. Either the amount of data in these classes was insufficient, or the shape homology with other classes was strong to explain these results.

The lack of data explained Precision and Recall of class 78. There were six available protein surfaces split into five protein surfaces in the training set, and one protein surface in the test set. This explanation was supported by empty classes \textit{i.e.} zero protein surfaces in these ground truth classes. Eight proposed methods predicted some protein surfaces in these classes, which can be explained by poor learning of data. But the lack of data cannot only explain Precision and Recall for class 34 (73 protein surfaces in the training set) and class 80 (57 protein surfaces in the training set). A majority of proposed methods displayed good success for classes of equivalent size (classes 51, 60 and 81) with stacked-Precision and stacked-Recall scores greater than 11.0 out of 15.0 (Supplementary Data: Figure 2).

Homology of classes 34 and 80 with others explained results of proposed methods. About 80\% of protein surfaces from these classes were not correctly classified in the ground truth classes, but predicted in homologous classes.

But, the proposed methods displayed also a majority of false negatives from ground truth classes either in non-homologous classes (\textit{e.g.} class 8) or evenly distributed between homologous and non-homologous classes (\textit{e.g.} class 14, class 56, class 88). These classes included between 121 and 818 protein surfaces in total. Local similarities~\cite{alcami2003, gainza2019deciphering, riahi2023_surfaceid, dokholyan2009} or local surfaces insufficiently detailed (Supplementary Data: Figure 5) can explain these false negatives in non-homologous classes. \\

\begin{figure}[htbp]
    \centering
    \includegraphics[width=0.5\linewidth]{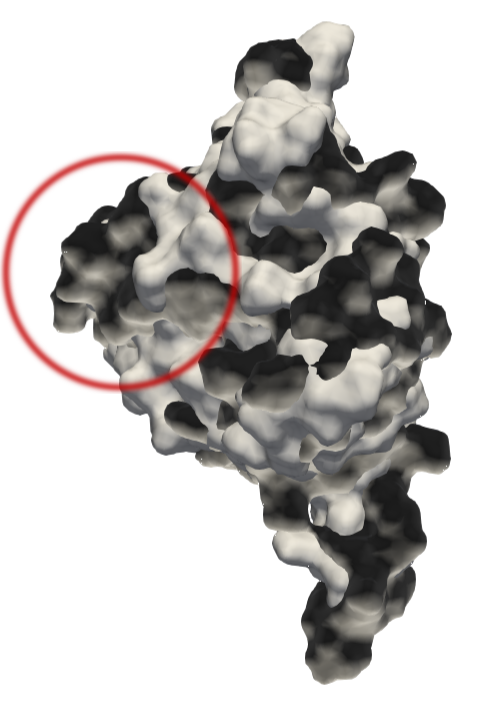}
    \caption{\textbf{Superposition of two protein surfaces from class 29.} In white: 6epc\_14:N:7 (from test set), in gray: 3unb\_13:AA:3 (from train set), in red: local dissimilarity due to an $\alpha$-helix. Due to local dissimilarities, 6epc\_14:N:7 was misclassified although the structural folding is globally similar to protein surfaces in training set (see Supplementary Data)}
    \label{fig:3Dsup}
\end{figure}

\noindent\textbf{Difficult protein surface classes.} 6epc\_14:N:7 and 6x5a\_4:H:H were not correctly predicted by all proposed methods. Three proposed methods classified 6epc\_14:N:7 in a non-homologous class, and thirteen proposed methods classified 6x5a\_4:H:H in a homologous class. To understand this result, we performed a careful visual inspection of these protein surfaces.

6epc\_14:N:7 belonged to the class 29. This class included 14 protein surfaces in the train set. The conformers displayed an C$\alpha$-based root mean square deviation of atomic positions (RMSD)  close to 0 between them, except with 6epc\_14:N:7. C$\alpha$-RMSD was 6\AA~between 6epc\_14:N:7 and the other conformers in its class. But, the folding and the global shape were similar visually (Figure \ref{fig:3Dsup}, Supplementary Data: Figure 4) in comparison with the other conformers. This RMSD was explained by a slight deviation of backbone locally, at the termini (N-terminal, C-terminal) or at a $\alpha$-helix region (delimited by residues from Alanine 131 to Methionine 147) for instance. This deviation caused significant local deformation of the surface, and explained the wrong classification by all proposed methods. 

6x5a\_4:H:H belonged to the class 34. This class included 73 protein surfaces in the train set. RMSD was lower than 1\AA between conformers of class 34. The folding and the global shape were similar between them. Conformers from this class were not always correctly classified. But, only 6x5a\_4:H:H were not correctly classified by all proposed methods. The success of prediction depends on the ability to discriminate local properties and/or local geometry. \\

\noindent\textbf{About using the electrostatic potential information} The proposed methods using the properties of electrostatic potential (\authorN{Barisin\_v2}{Author\_2\_v2}, \authorN{Tatsuma}{Author\_1} methods, 3D-SURFER) displayed the best performance for all classification metrics. This highlights the importance of this feature to improve model performance.

\authorN{Barisin \textit{et al.}}{Author\_2}'s participating team proposed two versions of their method. The first version used only the geometrical information of protein surfaces. The second version additionally used the electrostatic potential information. Accuracy, F1 score, Precision and Recall increased by 1\% and balanced accuracy increased by 5\% when the electrostatic potential was used. Even though more comparisons such as this one would be necessary to get stronger statistics on this point, the results from this group highlighted that using the electrostatic potential in addition to geometrical information can improve the performance in retrieval. 

Electrostatic potential influenced the retrieval performance of their method. Recall score increased for 25 classes and decreased for 11 classes. Precision score increased for 35 classes and decreased for 10 classes. F1 score increased for 36 classes and decreased for 15 classes. For class 23 (12 protein surfaces), class 29 (17 protein surfaces), class 57 (11 protein surfaces) and class 89 (4 protein surfaces), Precision increased from 0 to 1. The use of electrostatic potential displayed therefore a positive impact on the retrieval performance, and helped for classes with limited data in the training set.

\section{Conclusion}\label{conc}

This SHREC 2025 challenge on protein shape classification including the electrostatic potential was a great success, with 9 participating teams and offering a total of 15 proposed methods. This track proposed a large and unbalanced dataset (close to biological reality), with 11,565 protein surfaces divided into 97 classes. Most proposed methods used only geometrical information and 3 out of the 15 took into account the electrostatic potential information in addition to geometry. The proposed methods were highly diverse in data representation (point clouds, images) and were mostly ML/DL based (12 out of the 15 proposed methods were deep-learning based (DL) methods, 2 were machine-learning based (ML) methods).

Most proposed methods achieved good performance in retrieval reflected by their accuracy (between 71\% and 93\%) and F1 score (between 66\% and 92\%). The best one combined a point-cloud representation with the use of rotation invariant deep neural networks. Some proposed methods even achieved to handle an essential aspect, the data imbalance, shown by the balanced accuracy around 90\%. 

The best performing methods (with about 90\% of accuracy) used the electrostatic potential information. They outperformed the proposed methods that did not take into account this additional information. As illustrated by the F1 score and Precision and Recall, the performance of the proposed methods that used this information was particularly improved on certain classes that were more sensitive to this parameter. The use of electrostatic potential can improve the performance notably for classes with limited training data.

An interesting perspective to the improvement of the methods is to perform an ablation study, consisting in removing a subpart of a model, in order to see its impact on the performance. If this latter remains unchanged, the subpart can be definitely removed and this allows to improve the training and inference times of the model by simplifying it.

Preliminary results displayed the importance of the proposed methods sensitivity to local regions of protein surfaces. Local (dis)similarities can impact the retrieval performance as illustrated by particular molecular surfaces of the dataset that were misclassified by all proposed methods. 

Since the methods that took into acccount the electrostatic potential displayed the best performance in retrieval, future \MakeUppercase{SHREC} tracks on protein shape retrieval could include additional physicochemical properties into the dataset to assess their impact on retrieval.

\section{Acknowledgements}

T. Yacoub and M. Montes are supported by the French Agence Nationale de la Recherche (ANR22-CE23-GRADIENT) and by the European Research Council Executive Agency under the research grant number 640283.
The work of M. Guerra, U. Fugacci, G. Palmieri, A. Ranieri and S. Biasotti is funded by the European Union - NextGenerationEU and by the Ministry of University and Research (MUR), National Recovery and Resilience Plan (NRRP), Mission 4, Component 2, Investment 1.5, projects “RAISE - Robotics and AI for Socio-economic Empowerment” (ECS00000035), the "National Center for HPC, Big Data and Quantum Computing" (CN0000013) and the "National Center for Sustainable Mobility". This study has also been carried out in the framework of the activities of NRRP CN MOST, whose financial support is gratefully acknowledged.
Y. Kagaya, J.H. Park and D. Kihara acknowledge the National Institutes of Health (R01GM133840, R01GM123055) and the National Science Foundation (CMMI1825941, MCB1925643, IIS2211598, DMS2151678, DBI2003635, and DBI2146026) to DK for the partial support of this work.
F. Heidar-Zadeh acknowledges financial support from the Natural Sciences and Engineering Research Council (NSERC) of Canada, Compute Ontario, Digital Alliance of Canada, and Centre for Advance Computing (CAC) at Queen's University. A. Tehrani acknowledges support from Queen's University School of Graduate Studies and Postdoctoral Affairs (SGPS) Doctoral Award. F. Meng acknowledges the Banting Postdoctoral Fellowship and DRI EDIA Champions Pilot Program funded by Compute Ontario and the Digital Alliance of Canada.

\section{Declaration of Interest statement}

The authors declare no conflict of interest.

\bibliographystyle{unsrt}

\end{document}